\def\year{2021}\relax
\documentclass[letterpaper]{article} 
\usepackage{aaai21}  
\usepackage{times}  
\usepackage{helvet} 
\usepackage{courier}  
\usepackage[hyphens]{url}  
\usepackage{graphicx} 
\usepackage{natbib}  
\usepackage{caption} 
\usepackage[switch]{lineno} 
\usepackage{tabularx}
\usepackage{bm}
\usepackage{amssymb}
\usepackage{booktabs}
\usepackage{url}
\usepackage{multicol}
\usepackage{color}
\usepackage{arydshln}
\usepackage{multirow}
\usepackage{amsmath} 
\usepackage{amsfonts}
\usepackage{subfigure}
\def\first{$^1$}
\def\second{$^2$}
\def\third{$^3$}
\def\forth{$^4$}

\title{Probing Product Description Generation via Posterior Distillation }
\author{
    Haolan Zhan,\textsuperscript{\rm 1,4}\thanks{Work done at Data Science Lab, JD.com.}  
    Hainan Zhang,\textsuperscript{\rm 2}{\thanks{Corresponding authors.}}
    Hongshen Chen,\textsuperscript{\rm 2}\footnotemark[2]
    Lei Shen,\textsuperscript{\rm 3,4} 
    Zhuoye Ding,\textsuperscript{\rm 2} \\ Yongjun Bao,\textsuperscript{\rm 2} Weipeng Yan,\textsuperscript{\rm 2} Yanyan Lan\textsuperscript{\rm 3,4}\footnotemark[2] \\ }
\affiliations{
   \first Institute of Software, Chinese Academy of Sciences, Beijing, China\\
  \second Data Science Lab, JD.com, Beijing, China \\
  \third Institute of Computing Technology, Chinese Academy of Sciences, Beijing, China \\
  \forth University of Chiense Academy of Sciences, Beijing, China\\
  zhanhaolan316@gmail.com, zhanghainan6@jd.com, ac@chenhongshen.com, \{shenlei17z, lanyanyan\}@ict.ac.cn
}

\begin{document}

\maketitle

\begin{abstract}


In product description generation (PDG), the user-cared aspect is critical for the recommendation system, which can not only improve user's experiences but also obtain more clicks. High-quality customer reviews can be considered as an ideal source to mine user-cared aspects. However, in reality, a large number of new products (known as long-tailed commodities) cannot gather sufficient amount of customer reviews, which brings a big challenge in the product description generation task. Existing works tend to generate the product description solely based on item information, i.e., product attributes or title words, which leads to tedious contents and cannot attract customers effectively. To tackle this problem, we propose an adaptive posterior network based on Transformer architecture that can utilize user-cared information from customer reviews. Specifically, we first extend the self-attentive Transformer encoder to encode product titles and attributes. Then, we apply an adaptive posterior distillation module to utilize useful review information, which integrates user-cared aspects to the generation process. Finally, we apply a Transformer-based decoding phase with copy mechanism to automatically generate the product description. Besides, we also collect a large-scare Chinese product description dataset to support our work and further research in this field. Experimental results show that our model is superior to traditional generative models in both automatic indicators and human evaluation.
\end{abstract}
\section{Introduction}
\begin{figure}[!t]
\centering
\includegraphics[width=0.99\linewidth]{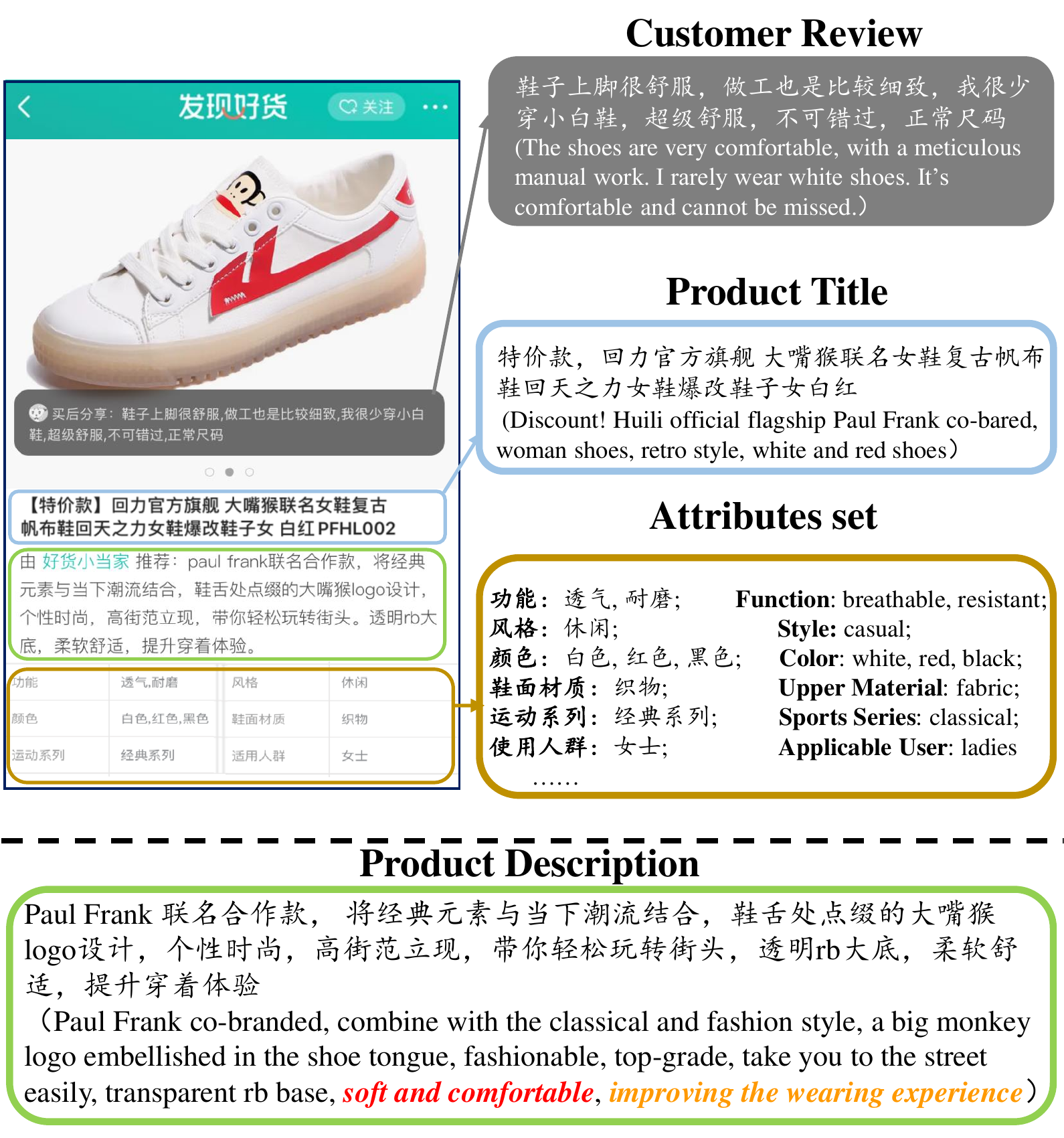}
\label{screenshot}
\caption{A product description example from our dataset.}
\label{fig:intro_case}
\end{figure}

In E-commerce, the goal of online product recommendation system is to post suitable commodities to customers and stimulate their purchasing behaviors. However, ranking the products and display them to users can no longer meet the requirements of customers \cite{zhang2019multi,gong2019automatic,chen2019towards}. While browsing the recommendation system, customers face the problem of information explosion. To save costs and find products in need straightforwardly, customers would like to see some refined product descriptions rather than complex product details, as shown in Figure \ref{fig:intro_case}. Therefore, it is important to present product characteristics in a product description for the E-commerce system to help customers learn the recommended products directly.

Furthermore, products with well-written description is capable to attract more customer's attentions. For instance, as showed in Figure \ref{fig:intro_case}, the product description below the product depicts some user-cared aspects, i.e., ``\textit{comfortable and soft}'' and ``\textit{improving the wearing experience}'', which can arouse customer's interests and encourage them to buy it. With this appealing user-cared description, customers could select their interested products easily and feel more satisfied with the entire recommending process. In a word, generating an user-cared product description is an important and practical research problem in E-commerce scenario. 

High-quality customer reviews are an ideal source to mine user-cared aspects \cite{pecar2018towards}. The customer post their reviews of a product, which naturally shows their most cared aspects. However, in reality, lots of new products (long-tailed products) cannot gather sufficient amount of customer reviews. We make some statistics of customer reviews in Table \ref{fig:reviews}. In category \textit{Shoes\&Clothes}, there are more than 66.3\% commodities have less than 10 reviews and the average number of reviews is only 18.4. The data shows that long-tailed phenomenon of customer reviews is obvious in the E-commerce system. That is to say, a large number of products lack enough corresponding reviews but we still need to generate product descriptions for them.

Recently, most existing methods \cite{zhang2019multi,li2020aspect} consider item contents, such as product image, text, attributes and title, as their source to generate the product description for long-tailed products. Obviously, the generated descriptions may be tedious and cannot attract customers effectively since they ignore user's experience. To enhance the effectiveness of user-cared aspects, other researchers \cite{chen2019towards,chen2019pog} propose to incorporate customer's personalized profiles and/or external product knowledge from Wikipedia to generate product descriptions. However, the personalized data is too sparse and thus hard to represent and utilize. On the other hand, these methods also cannot deal with the long-tailed products because the personalized data is inaccessible.

To tackle this problem, we propose an \textbf{A}daptive \textbf{P}osterior \textbf{D}istillation model based on \textbf{T}ransformer architecture (APDT), which can utilize user-cared aspects from customer reviews, and then incorporate these aspects into the generation process of product descriptions. Specifically, we first extend the self-attentive Transformer encoder to encode product items (title and attributes) and reviews. Then, we apply an adaptive posterior distillation layer to utilize effective review information. In this layer, product title and attributes representation are fused into item representation through feature fusion module at first. Then, the review representation is updated by interacted with item representation. During training phase, item and review representations are sent into decoder layer separately. KL divergence loss is employed in the distillation process to approximate item and review representations. Finally, we apply a Transformer decoding phase with copy mechanism to automatically generate product descriptions. Besides, to enhance the coherence between generated description and ground truth, we also employ a coherence-enhanced function during training.

In our experiments, to evaluate our automatic product description generation task, we construct a new Chinese dataset from JD.com, one of the biggest e-commerce platform in China. This dataset contains 345,799 pairs of item content and description. The  results on this dataset show that our model  outperforms the state-of-the-art generative baselines, in terms of both automatic and human evaluations. 


Our contributions are listed below: 1) We propose an adaptive posterior distillation Transformer model to tackle the long-tailed commodities problem in product description generation task. 2) We collect a large-scale Chinese product description dataset for this research point. 3) Experimental results on this dataset validate the effectiveness of our proposed model.

\begin{table}[t!]
    \centering
    \begin{tabular}{cccc}
    \toprule
     Category   & \#Products  & \#Review(Avg) & \#$<$10 \\ \hline
      Shoes\&Clothes   & 143,941  & 18.4 &  66.3\% \\
      Digital & 108,236  & 15.7 &  58.7\%  \\
      Homing  & 93,622  & 21.6 &  68.2\%  \\ 
    \bottomrule
    \end{tabular}
    \caption{Statistics of customer review information.}
    \label{fig:reviews}
\end{table}

\section{Proposed Method}

\begin{figure*}[!t]
\centering
\includegraphics[width=0.995\linewidth]{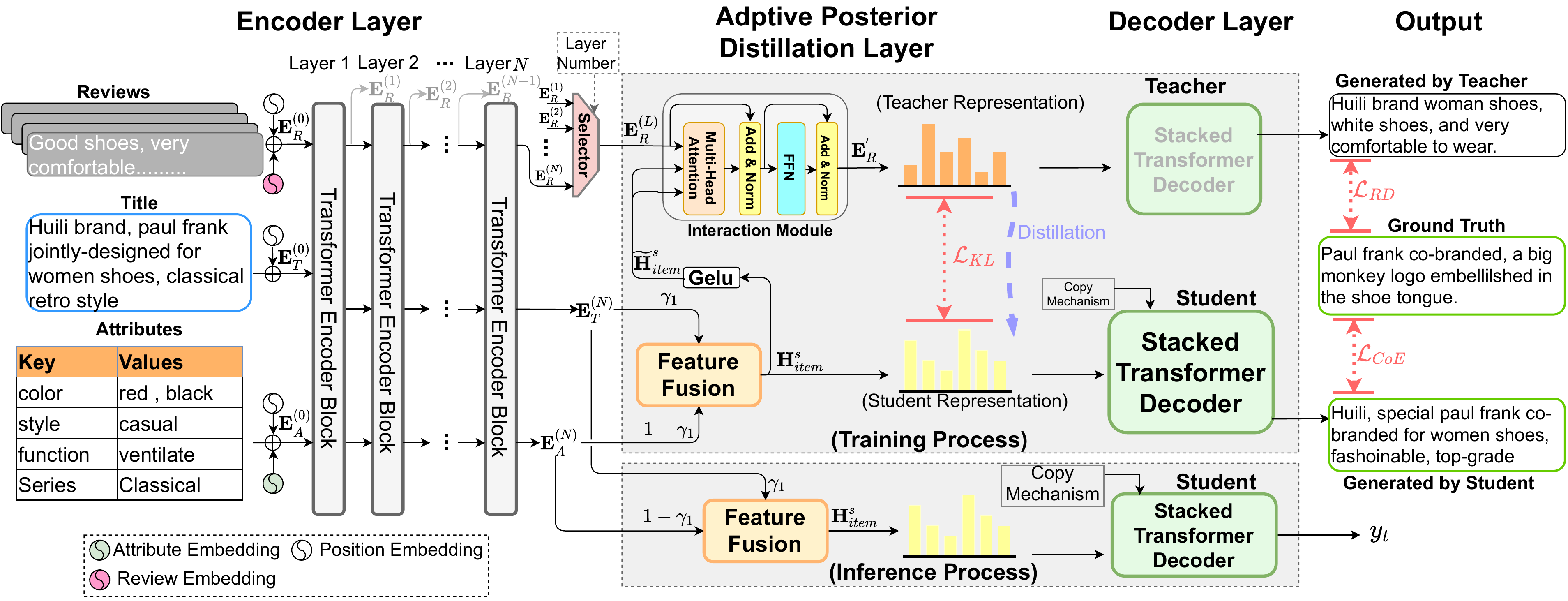}
\label{screenshot}
\caption{\small{ The architecture of APDT model. Here 
we omit the architecture inside the "Transformer Encoder Block" and "Stacked Transformer Decoder", and refer the readers to~\citet{vaswani2017attention} for more details.}}
\label{fig:model}
\end{figure*}

\subsection{Problem Formulation}

Given a product title, attribute sets and its corresponding customer reviews sets, the product description generation (PRG)  task aims at utilizing inherent product information at first,  and then identifying the privilege of customer's preference. Finally, coherent and appealing product descriptions will be generated.

Formally, given a product title $T$ composed of a sequence of words $\{t_1,...,t_N\}$, a group of attributes $A$ contains $M$ pairs of key-values, i.e., $A = \{\{k_1,a_1\},...,\{k_M,a_M\}\}$ and a relative customer review set defined as $\{R_1,...,R_n\}$, the PDG task attempts to learn a generative model $G(\cdot)$. Each review sentence in $R$ is defined as $R_n=\{r_n^1,\cdots,r_n^L\}$, where $r_n^i$ is the $i$-th word in the sentence of $R_n$ and $L$ is the max length for review sentence.  The corresponding generated product description is defined as $Y=\{y^1,\cdots,y^S\}$, where $y^i$ is the $i$-th word and $S$ is the max length of product description. With sequence-to-sequence framework, this can be formulated as follows:
\begin{equation*}
   P(Y | T, A, R; \theta) = \prod_{t=1}^{|S|} P(y_t|y_{<t}, T, A, \{R_{i}\}_{i=1}^{n}; \theta),
\end{equation*}
where $\theta$ is the parameter, and $y_{<t}$ denotes the previously generated words.

\subsection{Overview of Our Model}

As shown in Figure.~\ref{fig:model}, our model consists of three main layers: (1) an encoding
layer, (2) an adaptive posterior distillation layer, and (3) a decoding layer with copy mechanism. The encoding layer employs a stacked Transformer encoder module~\cite{vaswani2017attention} to encode context including attributes, titles and customer reviews. The adaptive posterior distillation layer contains feature fusion module, interaction module, and the teacher representation and student representation learning module. The decoding layer contains a stacked Transformer decoder to generate response $y_t$ token  by token.

\subsection{Encoding Layer}

In the encoding layer, we apply the Transformer encoder module with different position encoding mechanism to the title, attributes and reviews, separately.

Given a product title $T = \{t_1,...,t_N\}$ as the input, the initial word embedding and  position embedding vectors are represented as $WE(T)$ and $PE(T)$ respectively. The initial input title representations  $\mathbf{E}_{T}^{(0)}$ is the sum of word and position embedding at the first layer:
\begin{equation*}
    \mathbf{E}_{T}^{(0)} = WE(T) + PE(T).
\end{equation*}

At the $l$-th layer, the output representation is defined as below: 
\begin{equation*}
    \mathbf{E}_{T}^{(l)} = {\rm FFN}({\rm MHA}(\mathbf{E}_{T}^{(l-1)}, \mathbf{E}_{T}^{(l-1)}, \mathbf{E}_{T}^{(l-1)})),
\end{equation*}
where $\mathbf{E}_{T}^{(l)}$ denotes the output representations after the $l$-th layer. The sub-layer $\rm FFN(\cdot)$ is a position-wise fully connected feed-forward network,  and ${\rm MHA}(\mathbf{Q,K,V})$~\cite{vaswani2017attention} is a multi-head attention function. We refer the readers to \citet{vaswani2017attention} for more details.

For attributes context, we apply a unique attribute embeddings (AE), to adapt to its structured data format. Attribute embeddings are employed to differentiate the key-value pairs in the attribute sets.
Therefore, inside the key-value pair, the words share same attribute embeddings. The initial representation of attributes representation  $\mathbf{E}_{A}^{(0)}$  and encoding phases are defined as:
\begin{align*}
   \mathbf{E}_{A}^{(0)} &= WE(A) + PE(A) + AE(A), \\
   \mathbf{E}_{A}^{(l)} &= {\rm FFN}({\rm MHA}(\mathbf{E}_{A}^{(l-1)}, \mathbf{E}_{A}^{(l-1)}, \mathbf{E}_{A}^{(l-1)})).
\end{align*}

For customer review representations, given a review set $\{R_1,...,R_N\}$ as the input, we firstly concatenation all the words as a sequence. Then, we will apply a review embeddings (RE) to differentiate the review sentences. The initial representation of review sentences  $\mathbf{E}_{R}^{(0)}$  and encoding phases are defined as:
\begin{align*}
        \mathbf{E}_{R}^{(0)} &= WE(R) + PE(R) + RE(R),\\
    \mathbf{E}_{R}^{(l)} &= {\rm FFN}({\rm MHA}(\mathbf{E}_{R}^{(l-1)}, \mathbf{E}_{R}^{(l-1)}, \mathbf{E}_{R}^{(l-1)})).
\end{align*}

\subsection{Adaptive Posterior Distillation Layer}

Inspired by the knowledge distillation's success~\cite{hinton2015distilling,feng2020regularizing} on model compression and knowledge transfer, we propose an adaptive posterior distillation layer to transfer user-cared aspects in review information (\textit{teacher}) to item representation (\textit{student}), which contains title and attributes information.

During the posterior training process, we design an individual training objective for reviews (\textit{teacher}) information,
in order to enhance the semantic coherence between review information and target product description.

\textbf{Student representation}: 
Firstly, we define an item representation to combine title and attributes representation.
The item representation (\textit{student}) is defined as:
\begin{equation*}
    \mathbf{H}^{s}_{item}=\gamma_{1}\mathbf{E}_{T}^{(N)}+(1-\gamma_{1})\mathbf{E}_{A}^{(N)},
\end{equation*}
where $\gamma_{1} \in [0,1]$ is a parameter, and $\mathbf{E}^{(N)}_{T}$ and $\mathbf{E}^{(N)}_{A}$ is the final representation of title and attributes that output from the $N$-th encoder layer, also known as the last encoder layer.

\textbf{Teacher representation}: Given the reivew representation $\mathbf{E}_{R}^{(N)}$, we firstly apply an interaction module to incorporate item information (title and attributes) into the representation of review information. The interaction module is designed to highlight user-cared aspects in reviews, with the assistance of item representation. It is a one layer mutli-head attention following with a feed-forward sub-layer.

To make all dimension of representation matrix compatible, we perform a non-linear projection of the parameters in student representation $\mathbf{H}^{s}_{item}$ before fed into interaction module. Therefore, the updated item representation (student) $ \widetilde{\mathbf{H}}^{s}_{item}$ and updated review representations (\textit{teacher}) $\mathbf{E}_{R}^{'}$ are defined as:
\begin{align*}
    & \widetilde{\mathbf{H}}^{s}_{item} = Gelu({W}_{1} \mathbf{H}^{s}_{item} + {b}_{1}),\\
    & \mathbf{E}_{R}^{'} = {\rm FFN}({\rm MHA}(\widetilde{\mathbf{H}}^{s}_{item}, \widetilde{\mathbf{H}}^{s}_{item}, \mathbf{E}_{R}^{(L)})),
\end{align*}
where $W_1$, $b_1$ are the parameters,  and $Gelu$ (Gaussian Error Linear Unit)~\cite{hendrycks2016gaussian} is the non-linear projection function. $\mathbf{E}_{R}^{(L)}$ is the $L$-th encoder layer output, and $L \in [1,N]$. $L$ is a variable and set by human in experiments. As shown in Figure~\ref{fig:model}, Selector module (pink) is used to select the $L$-th encoder layer output.


we apply a stacked Transformer decoder layer to output $p^{r}_{l}(y_{t})$ as the  probability of token $y_{t}$  generated by
the teacher model at the $t$-th step. 
Therefore, the review distillation training objective $\mathcal{L}_{RD}(\theta)$ as:
\begin{equation*}
   \mathcal{L}_{RD}(\theta) = -\sum_{t=1}^{S} {\rm log} p^{r}_{l}(y_t|\mathbf{E}_{R}^{'};\theta)
\end{equation*}
where $p^{r}_{l}(y_{t})$ is the representations generated by the $\mathbf{E}_{R}^{'}$ respectively for a token $y_{t}$. 

During the posterior training phase, in order to approximate the distributions of student and teacher representation, we introduce the KL divergence loss, to measure the proximity between the prior item (student) representation and the posterior review (student) representation. The KL-divergence is defined as follows:
\begin{equation*}
   \mathcal{L}_{KL}(\theta)= D_{KL}(p^{s}({y_{t}}| {\mathbf{H}}^{s}_{item}) || p^{r}_{l}({y_{t}}|\mathbf{E}_{R}^{'});\theta),
\end{equation*}
where $\theta$ denotes the model parameters.

In the inference process, we only keep the well-trained prior module, and then feed the item representation  ${\mathbf{H}}^{s}_{item}$ into decoder layer.

\subsection{Decoding Layer}

In the decoding layer, we apply a stacked Transformer decoder module equipped with a copying mechanism~\cite{see2017get} to generate product description. We feed the product representation  ${\mathbf{H}}^{s}_{item}$ into decoder layer. Specifically, the probability of generating token $y_t$ at $t$-th step is modeled as:
\begin{equation*}
   P(y_t) = \lambda_{1}P_{vocab}(y_t|{\mathbf{H}}^{s}_{item}) + \lambda_{2}P_{cp}(y_t|\mathbf{E}_{T}) + \lambda_{3}P_{cp}(y_t|\mathbf{E}_{A})
\end{equation*}
where $P_{cp}(y_t|\mathbf{E}_{T})$ derives the copying probability from title words. The copy mechanism is defined as follows:
\begin{equation*}
  P_{cp}(y_t|\mathbf{E}_{T}) = \sum_{i:t_{i}=y_t}\alpha_{t,i},
\end{equation*}
and $P_{cp}(y_t|\mathbf{E}_{A})$ derives the copying probability from attributes words, which is calculated in a similar way. $P_{vocab}(y_t|{\mathbf{H}}^{s}_{item})$ is the output probability from a stack of Transformer decoder layers~\cite{vaswani2017attention}. $\lambda_{1}$, $\lambda_{2}$ and $\lambda_{3}$ are the coordination probability, which are estimated as follows:
\begin{equation*}
   [\lambda_{1},\lambda_{2},\lambda_{3}] = softmax({W}_{2}{\mathbf{H}}^{s}_{item} + {W}_{3}\mathbf{E}_{T} + {W}_{4}\mathbf{E}_{A} + {b}_{2}),
\end{equation*}
where ${W}_{2}, {W}_{3},{W}_{4},{b}_{2}$ are the parameters.


\subsection{Training Objectives}


Besides applying the KL-divergence loss function for posterior distillation module, we also employ a \textbf{Co}herence-\textbf{E}nhanced  Negative Log-Likelihood objective (CoE), which aims to help our model to generate words that seldom mentioned but coherent to user-cared aspects.

Different commodities which describe different aspects are always featured by the unique attribute values in the dataset. For example, a \textit{clothes} category often has the attributes like 'texture', 'size'. The information in the unique attributes is harder to capture than that in the common attributes like 'name', as the latter attributes are very frequent in the training set. We define the frequency of an attribute word $a_{k}$ as $f(a_k) = [freq(a_k)]^{-1}$ by calculating its frequency in the training set.

For a generated description $y^{\star}$, the coherence score between $y^{\star}$ and ground truth $y_{g}$ is calculated as follows:
\begin{equation*}
    Coh(y^{\star}) = \frac{\sum_{i=1}^{|y^{\star}|}(f(y^{\star}_{i})\cdot \mathbb{B}\{y^{\star}_{i} \in y_{g}\} )}{\sum_{i=1}^{|y|}f(y_{i})},
\end{equation*}
where $f(\cdot)$ is the word frequency index, and $\mathbb{B}\{y^{\star}_{i} \in y_{g}\} = 1$ if word $y^{\star}_{i}$ is in the ground truth sentence $y$. If not, it equals to $0$.

\noindent \textbf{Coherence-enhanced Function}: Different from previous models which only measures how well the generated sentences match the target sentences, we design a fused coherence-enhanced function $R_{fuse}$ which contains both the information coherence score and the ROUGE-L score (RG for short) of the generated descriptions. 
\begin{equation*}
    R_{fuse}(y^{\star}) = \beta R_{Coh}(y^{\star}) + (1-\beta)R_{RG}(y^{\star}),
\end{equation*}
where $\beta$ is set to 0.4. $R_{Coh}(y^{\star})$ is the coherence score between $y^{\star}$ and $y_{g}$, while $R_{RG}(y^{\star})$ is designed to calculate the ROUGE score.

We apply a coherence-enhanced negative log-likelihood (CoE) as our training objective The training loss of the generation task is defined as:
\begin{equation*}
    \mathcal{L}_{CoE}(\theta) = -\sum_{t=1}^{|S|} {R_{fuse}(y_{<t})}\cdot{log}(p(y_t|y_{<t}, T, A;\theta)),
\end{equation*}
Therefore, we optimize our all the following objectives jointly:
\begin{equation*}
   \mathcal{L}_{all}(\theta) = \alpha\mathcal{L}_{CoE}(\theta) + (1-\alpha) \bigg( \mathcal{L}_{RD}(\theta) + \mathcal{L}_{KL}(\theta) \bigg ),
\end{equation*}
where $\alpha \in [0,1]$, and it is used to weigh the contribution of different losses. A high value
of $\alpha$ makes the student model focus more on generation task; whereas a relative lower value of
$\alpha$ makes the student learn more from the teacher.


\section{Experiments}

\subsection{Experimental Settings}

\subsubsection{Dataset} We collect a large-scale Chinese product description generation dataset, named as JDPDG from  JD.com\footnote{\url{https://www.jd.com/}}, one of the biggest e-commerce platforms in China. Our dataset contains over 300 thousands product instances from the \textit{Clothes\&Shoes}, \textit{Digital} and \textit{Homing} categories. There are 104 kinds of products in \textit{Clothes\&Shoes} category,such as T-shirts and boots; 79 kinds of products in Digital, such as cameras and phones; 96 kinds of products in Homing, such as bowls and tobacco jars. Each commodity instance in our dataset includes a set of product information and a well-written product description. The set of product information contains a title, a group of attributes and a set of customer reviews. The product descriptions are written by thousands of qualified writers, with the reference of product title and attributes. The review information will be filtered at first, and only the the high-quality reviews are kept. The average number of words in each title, review and product description sentence are 13.8, 25.6 and 40.2, respectively. The average number of attribute keys in each product is 9.5, and for each key, its corresponding value contains 1 to 4 words. Table~\ref{fig:datasets} shows more details about our dataset\footnote{https://github.com/jddsl/JDPDG}.

\begin{table}[t!]
    \centering
    \begin{tabular}{lccc}
    \toprule
     Category & \textit{Shoes\&Clothes} & \textit{Digital} & \textit{Homing} \\ \hline
     Training Pairs   & 135,941  & 100,236 & 85,622 \\
     Validation Pairs  & 4000 &  4000 & 4000  \\
     Test Pairs   & 4000 &  4000 & 4000  \\
    \bottomrule
    \end{tabular}
    \caption{Data statistics for our proposed JDPDG dataset.}
    \label{fig:datasets}
\end{table}

\subsubsection{Baseline Models} We compare our adaptive posterior distillation Transformer (APDT) model with several baseline models, including:  
(i) \textbf{PG-BiLSTM}: a bi-directional LSTM with pointer generator mechanism \cite{see2017get}, 
(ii) \textbf{MS-Ptr}: a multi-source pointer network for short product title generation \cite{sun2018multi},
(iii) \textbf{Transformer}: an encoder-decoder
architecture relying solely on self-attention mechanisms~\cite{vaswani2017attention},
(iv) \textbf{HierTrans}: a hierarchical transformer for abstractive multi-document summarization tasks \cite{liu-lapata-2019-hierarchical},
(v) \textbf{EMA}: a unified text generation model for both structured and unstructured data with exponential moving average (EMA) technique \cite{shahidi-etal-2020-two},
(vi) \textbf{KOBE}: the state-of-the-art product description generation model with incorporated personalized knowledge attributes from external Wikipedia knowledge base~\cite{chen2019towards}. 


\subsubsection{Evaluation Metrics} We conduct both automatic and human evaluations. For automatic evaluation we follow previous PDG studies and use BLEU~\cite{papineni2002bleu} and ROUGE-L~\cite{lin-2004-rouge}.  For human evaluation we randomly sample 200 examples from each test set. For each example, we ask six workers (both CS graduate students) to conduct a pairwise comparison between the product description generated by our APDT and other baselines. Specifically, each worker needs to give a preference in terms of three
criteria: (1) Correctness, i.e., which description contains most correct information; (2) Diversity, i.e., which description looks more diversity;
(3) Coherence, i.e., which description looks mostly coherent to the product. Each criterion is assessed
with a score range from 1 (worst) to 4 (best).

\begin{table*}[!t]
\small
\centering
\begin{tabular}{l|cc|cc|cc}
\toprule[1pt]
  \multicolumn{1}{c|}{}  & \multicolumn{2}{c|}{\textit{Clothes\&Shoes}} & \multicolumn{2}{c|}{\textit{Digital}} & \multicolumn{2}{c}{\textit{Homing}} \\ 
  Model & ROUGE-L & BLEU & ROUGE-L &  BLEU & ROUGE-L &  BLEU\\
  \hline
  PG-BiLSTM~\cite{see2017get} & 15.62 & 7.86 & 16.86 & 8.02 & 15.17 & 7.59   \\
  MS-Ptr~\cite{sun2018multi}  & 15.95 & 7.98 & 16.54 & 7.79 & 15.72 & 7.74  \\ 
  Transformer~\cite{vaswani2017attention}  & 16.38 & 7.83 & 16.64 & 7.63 & 16.58 & 7.31\\
  HierTrans~\cite{liu-lapata-2019-hierarchical} & 17.36 & 8.51 & 17.73 & 8.46 & 16.89 & 8.28  \\
  EMA~\cite{shahidi-etal-2020-two}  & 16.32 & 8.69 & 17.67 & 8.81 & 16.53 & 9.33   \\
  KOBE~\cite{chen2019towards}  & 19.07 & 9.27 & 18.97 & 9.32 & 18.72 & 9.41 \\ \hline
  \textbf{APDT} ({ours})&  \textbf{20.41} & \textbf{10.36} & \textbf{19.95} & \textbf{10.08} & \textbf{19.68} & \textbf{10.13}  \\
 \bottomrule[1pt]
\end{tabular}
\caption{Automatic evaluation results on PDG dataset, including three different categories (\%).}
\label{tab_auto-eval}
\end{table*}

\subsubsection{Implementation Details}
We implement our model in OpenNMT\footnote{\url{https://github.com/OpenNMT/OpenNMT-py}} and train all models on the Tesla P40 GPUs with Pytorch \cite{paszke2019pytorch}. For experimental models, the hidden units of all transformer-based models are set as 512 and the feed-forward hidden size is set as 1,024. The beam search size is set as 5 and length penalty as $\alpha$ = 0.4~\cite{wu2016google}. For LSTM-based models, the word dimension is set to 300 and the hidden nodes are set as 256 for the encoder and decoder. The dropout rate and smoothing factor are set as 0.1 \cite{fabbri-etal-2019-multi}. The initial learning rate is set to 0.001. The $\beta_1$ = 0.9 and $\beta_2$ = 0.998 are used for gradient optimization. We also apply warm-up trick over the first 8, 000 steps, and decay as in \citet{vaswani2017attention}.  For hyper-parameters, we set $\gamma_{1}$, $\beta$ and $\alpha$ to 0.5, 0.4 and 0.5, respectively. 

\subsection{Experimental Results}

\subsubsection{Automatic Evaluation}
The automatic evaluation results are shown in Table~\ref{tab_auto-eval}. Our proposed APDT model outperforms the best. Taking the ROUGE metrics as an example, the ROUGE-L value of the APDT in the \textit{Clothes\&Shoes} category is 20.41, which is significantly better than MS-Ptr, HierTrans, EMA and KOBE models i.e., 15.95, 17.36, 16.32 and 19.07. The BLEU metrics of our model is also higher than other baseline models,indicating that our model can generate more informative and fluency product description. We also conducted a significant test, showing that the improvement is significant, i.e., p-value $<$ 0.01.



\begin{table}[!h]
\small
\centering
\begin{tabular}{l|ccc}
\toprule[1pt]
  Model & Correctness & Diversity & Coherence \\
  \hline
  PG-BiLSTM & 2.58 & \textbf{3.29} & 2.62  \\
  MS-Ptr & 2.62 & 3.26 & 2.57 \\ 
  Transformer & 2.19 & 2.87 & 2.30  \\
  HierTrans & 2.45 & 3.04 & 2.54   \\
  EMA & 2.63 & 2.91 & 2.75\\
  KOBE & 2.47 & 3.11 & 2.91  \\\hline
  \textbf{APDT} (ours) & \textbf{2.91} & {3.27} & \textbf{3.02} \\
 \bottomrule[1pt]
\end{tabular}
\caption{Human evaluation on clothes\&shoes category.}
\label{tab_human_clothes}
\end{table}

\subsubsection{Human Evaluation} 
We further conduct human evaluations to assess the proposed APDT model. Due to the limitation of pages, we only present the evaluation results on \textit{clothes\&shoes} category. But results on other two categories also show a similar trend. 
Table~\ref{tab_human_clothes} summarizes the evaluation results. In the correctness criterion, our APDT model achieves a scores at 2.91, while other baseline models only get scores about 2.5. This result indicate that our model can generate more correct aspects. In the coherence criterion, APDT model can also achieves the best performance, indicating that APDT model can generate coherent and relevant information than baselines.
We also employ Fleiss’ kappa scores~\cite{fleiss1971measuring} to measure the reliability between different annotators. The overall Fleiss' kappa score is 0.527.


\begin{table}[t]
\centering
\begin{tabular}{l|cc}
\toprule[1pt]
  Model & ROUGE-L & BLEU  \\
  \hline
  \textbf{APDT} &  \textbf{20.41} & \textbf{10.36}   \\ \hline
  \quad -- \textit{Copy Mechanism} & 19.13 & 9.75   \\
  \quad -- \textit{Posterior Distillation} & 18.56 & 9.79   \\
  \quad -- \textit{Coherence Enhanced} & 19.87 & 10.02   \\
  \quad -- \textit{above all} & 17.95 & 9.13   \\
 \bottomrule[1pt]
\end{tabular}
\caption{Ablation test on the clothes\&shoes category (\%).}
\label{tab_ablation}
\end{table}

\subsection{Case Study}

\begin{figure*}[!t]
\centering
\includegraphics[width=1.0\linewidth]{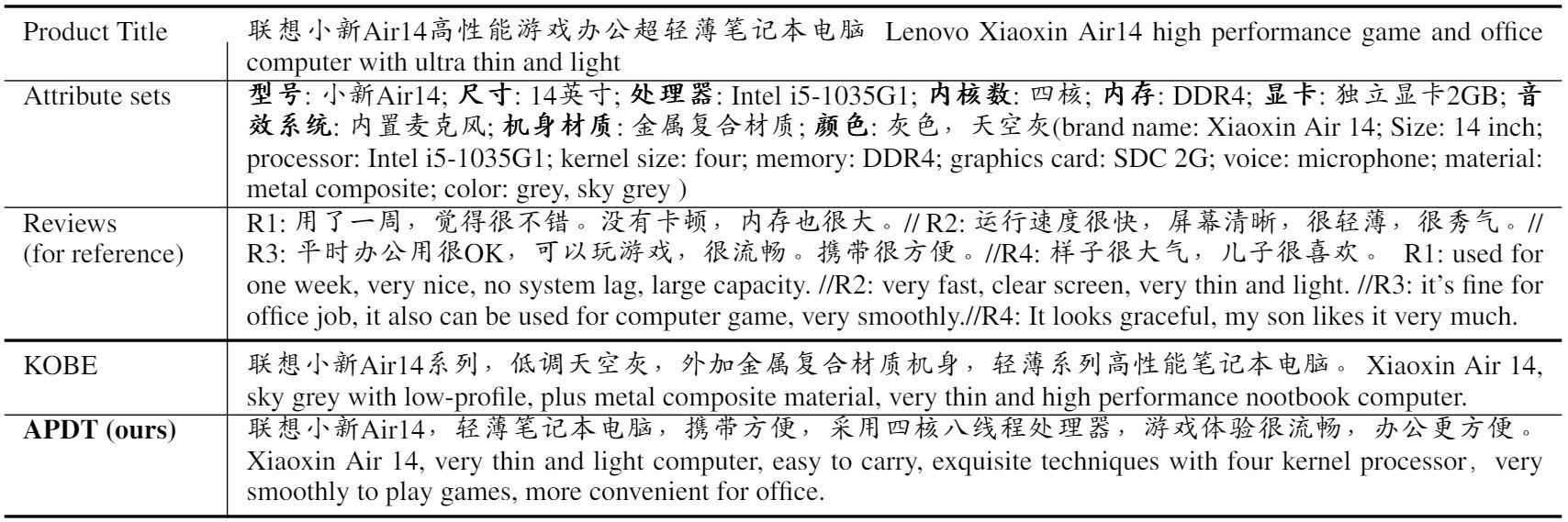}
\caption{\small{Case study of APDT and baselines on JDPDG dataset.}}
\label{tab_finalcase}
\end{figure*}

To facilitate a better understanding of our model, we present some examples in Table \ref{tab_finalcase}. With the page limitation, we only present the generated production description from KOBE and our APDT model. For fair comparison, during inference process, we only send the product title and attributes sets into these two models. Review information are presented only for reference. As shown in Table~\ref{tab_finalcase}, our proposed APDT model generates more aspects of product with considering customer review information. For example, the product description generated by KOBE model can cover the relevant and appropriate information in product title and attributes, such as ``
\textit{Xiaoxin Air 14}'', and ``\textit{sky grey}''. However, it's may difficult for KOBE model to generate user-cared aspects without the assistance of our proposed posterior distillation module. Our proposed APDT model is able to contain more user-cared information, such as ``\textit{easy to carry}'', ``\textit{very convenient for office}'' and ``\textit{very smoothly}'', since it has learned from the distillation information from reviews (teacher) representation during posterior training phase.

\subsection{Model Analysis}


\subsubsection{Effect of the Copy Mechanism}
To include as many relevant and correctness aspects in the generated product description, the proposed APDT model involves a copy mechanism during the decoder phase. We ablate the copy mechanism from the framework
by using only naive transformer decoder to verify its effectiveness. As showed in Table~\ref{tab_ablation}, 
we can witness that the absence of copy mechanism hurts performance of APDT model. The ROUGE-L and BLEU scores decrease from 20.41 to 19.13, and 10.36 to 9.75, respectively. 
It demonstrates that the copy mechanism plays an important role in achieving strong performance.

\begin{figure}[t]
\centering
\includegraphics[width=0.65\linewidth]{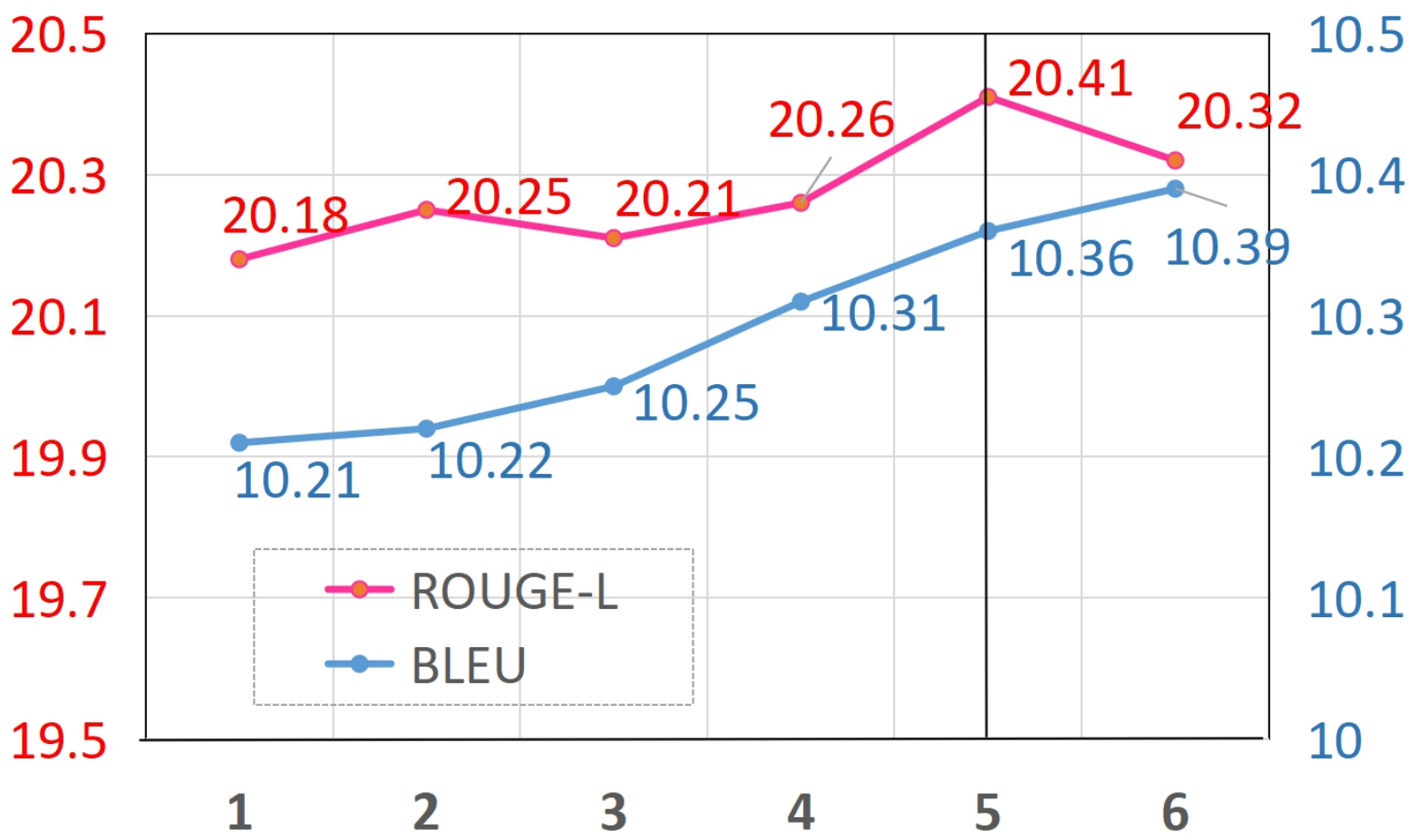}
\label{screenshot}
\caption{\small{The ROUGE-L (red) and BLEU (blue) scores (\%) with different encoder layer of review (teacher) representation in the posterior distillation module.}}
\label{fig:dist-layer}
\end{figure}

\begin{figure}[!t]
\centering

\subfigure[ROUGE-L score (\%).]{
\begin{minipage}[t]{0.47\linewidth}
\centering
\includegraphics[width=1\linewidth]{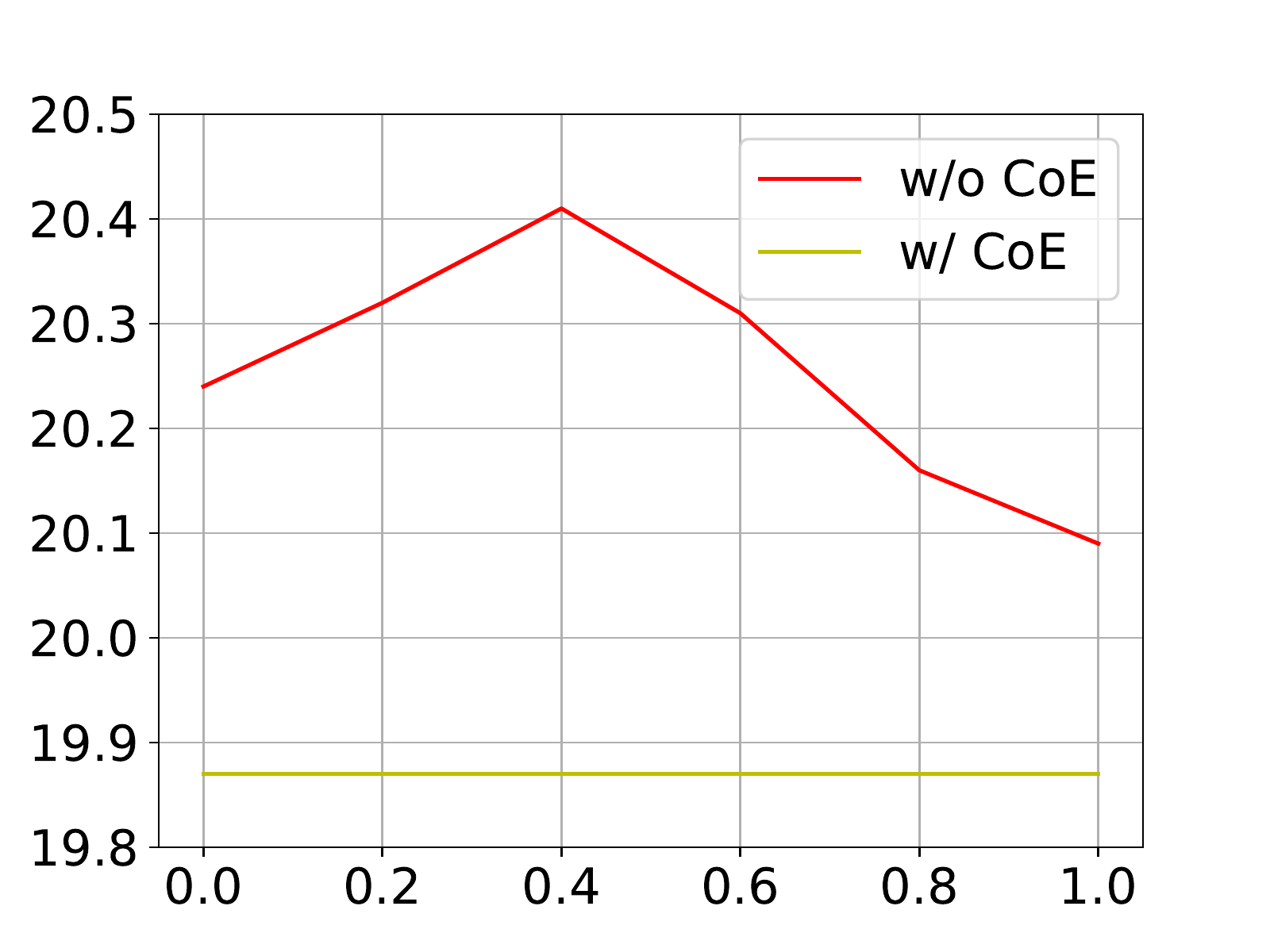}
\label{b-rouge}
\end{minipage}
}
\subfigure[BLEU score (\%).]{
\begin{minipage}[t]{0.47\linewidth}
\centering
\includegraphics[width=1\linewidth]{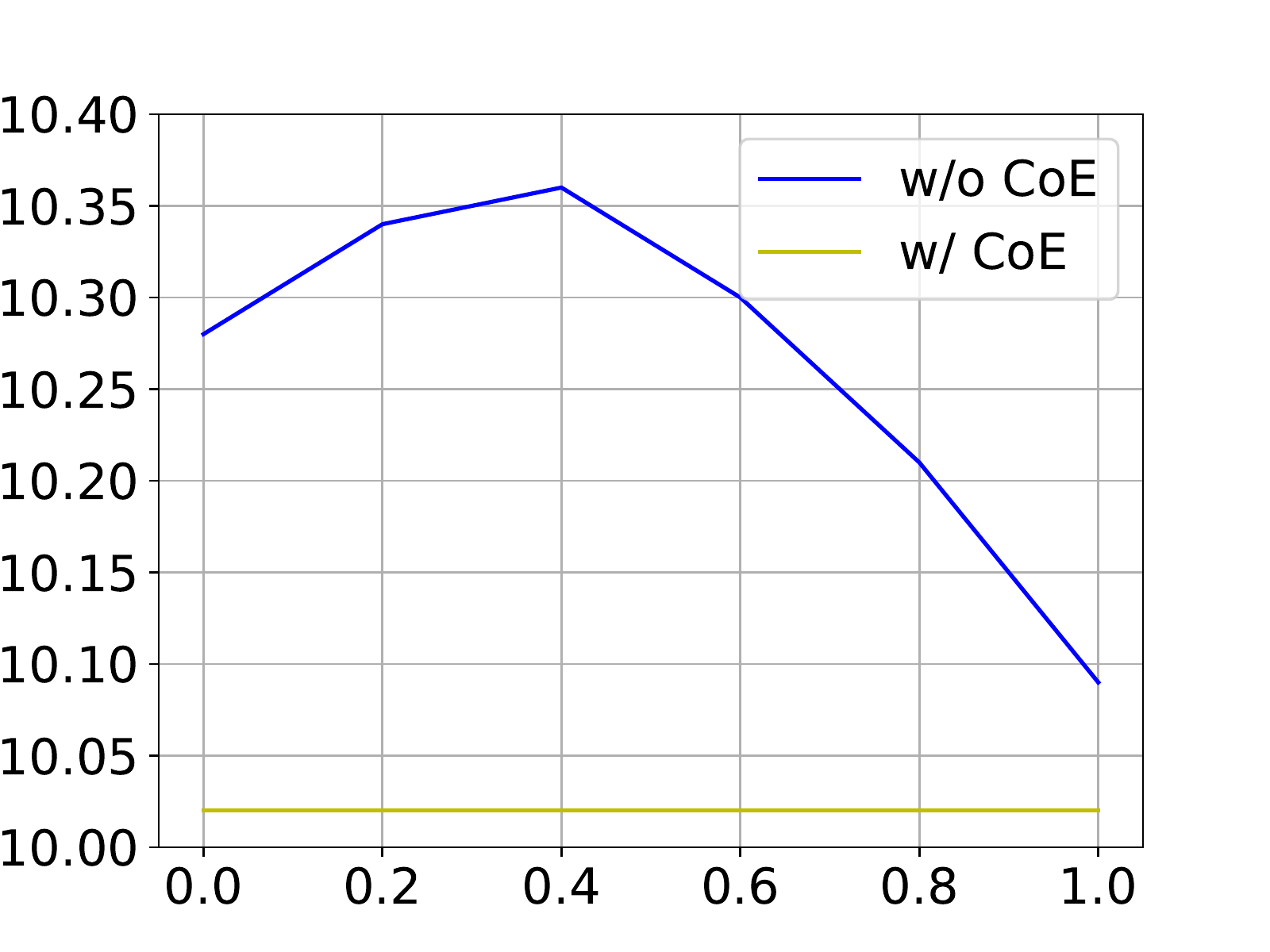}
\label{b-bleu}
\end{minipage}
}%

\centering
\caption{\small{the analysis of $\beta$ in the coherence enhanced function on the validation dataset. Red line in (a) represents the ROUGE-L socre of model with CoE loss, while blue line in (b) stands for the BLEU score of model with CoE loss. Yellow line in both (a) and (b) represent the model without CoE loss.}}
\label{fig:b-choose}
\end{figure}

\subsubsection{Effect of the Posterior Distillation}
In our proposed APDT model, posterior distillation can distill user-cared aspect information from review
information, allowing the student model to generate the user-cared aspects in the description for long-tailed commodities. From Table~\ref{tab_auto-eval}, the ablating of posterior distillation also leads to a performance drop in the automatic evaluation metrics.

Furthermore, to analyze the distillation effects on product (student) representation, we conduct an experiment to identify which encoder layer that review (teacher) representation distill from.  From Figure~\ref{fig:dist-layer}, we observe that the product (student) representation benefits the most from distilling the $5$-th or the $6$-th encoder layer of the review representation (teacher). In the shallow encoder layer, student representation may not be able to learn well from the user's aspects information. On the other hand, it's easier for student model to mimic from a teacher representation that comes from a deeper layer.

\subsubsection{Effect of the CoE Loss}
To evaluate the performance of coherence enhanced negative log-likelihood loss, we explore the impact of using different $\beta$ value in the coherence-enhanced function. As presented in Figure~\ref{fig:b-choose}, we observe that when $\beta < 0.4$, the increasing of $\beta$ leads to continuous improvement on the BLEU and ROUGE-L metric. However, the performance of our model tends to decrease when $\beta$  continues rising. Therefore, overemphasis on the coherence-enhanced function will finally risk introducing inconsistent aspects information into the final description sentences.


\section{Related Work}



\subsection{Text Generation in E-commerce}


Text generation in E-commerce aims at improving customer's online shopping experience. Several novel and challenging tasks are proposed, including short title generation~\cite{zhang2019multi}, product description generation~\cite{chen2019towards} and recommendation reason generation~\cite{zhan2020user}. The motivation of STG is to concisely display short product titles on limited screen of mobile phones.  \citet{gong2019automatic} firstly proposed the short title generation task for e-commerce, which automatically generates short title by directly extracting essential information from original long title. 
\citet{wang2018multi} proposed a multi-task learning approach by using external searching log data as additional task to facilitate key words extraction process. Furthermore, \citet{zhang2019multi} considered a multi-source approach incorporating multi-modal information with generative adversarial networks.
As for product description generation task, early work focuses on template-based generation approaches that incorporates statistical methods~\cite{wang2017statistical}. With the evolution of neural network methods, RNN and Transformer are applied in this task.  
\citet{chen2019towards} proposed a personalized knowledge transformer model to generate the product description. Their methods utilized the item-based features, i.e., product image, attributes and title, and external knowledge base, such as Wikipedia. However, the external knowledge base 
risks introducing noise, which may hurt the effectiveness of generating personalized product description. 

\subsection{Personalized Content Generation}

Personalized content generation has attracted research interest in various domains, e.g., E-commerce~\cite{zhao2019dynamic,chen2019fine}, the automatic generation of marketing messages \cite{roy2015automated,chenlrec2020jddc}, persuasive message \cite{ding2016personalized,zhang2018speeding}, poetry generation~\cite{shen2020compose}, argument generation \cite{carenini2006generating} and dialogue generation~\cite{shen2020cdl,feng20aaai,shen2019modeling,shen2021learning,hengyi2020ijcai,liu2020nlpcc}. With the support of user preferences, the effectiveness has increases. Recently, \citet{krishna2018vocabulary} presented a framework for the summary generation that takes into consideration the linguistic preferences of the specific audience.
\citet{reichelt2014talk} showed that personalized information of learning materials can increase motivation and learning outcomes. \citet{zander2015does} studied the effect of personalization on students' attention allocation using some eye-tracking methods, and find that the personalized parts of reading materials are more attractive. In the field of E-commerce, 
\citet{elad2019learning} proposed an extractive method to select sentences and then generate personalized product description. \citet{chen2019pog} built a bridge between personalized outfit generation and recommendation by considering both user preferences and individual items. 
To the best of our knowledge, our method takes the first attempt to introduce user-cared aspects for product generation task.

\section{Conclusion}


In this paper, we propose an adaptive posterior distillation method for product description generation task. This method enables our Transformer-based model to utilize customer reviews and incorporate user-cared aspects into product description, especially for the long-tailed commodities. To better evaluate our proposed approach, we also construct a new Chinese product description dataset CPDG, and then present an adaptive posterior distillation method, which can distill user-cared aspects to the product description generation process. Extensive experiments conducted on our proposed dataset show that our proposed method could achieve better performance than baseline models. In future work, we plan to further investigate the proposed model with question-answering information, and then extend our approach to a multi-task framework, which is capable to handle a joint user intent recognition task.

\section{Acknowledgements}

The authors would like to thank Hengyi Cai from Institute of Computing Technology, Chinese Academy of Sciences, and the anonymous reviewers for their constructive comments and suggestions. This work was partially supported by the National Key R\&D Program of China under Grants No. 2019AAA0105200, 2016QY02D0405, the Beijing Academy of Artificial Intelligence (BAAI) (No. BAAI2020ZJ0303), the National Natural Science Foundation of China (NSFC) (No. 61722211, 61773362, 61872338, 61906180).

\bibliography{reference1}

\end{document}